\pdfoutput=1

\documentclass[11pt]{article}

\usepackage[]{acl}

\usepackage{times}
\usepackage{latexsym}

\usepackage[T1]{fontenc}

\usepackage[utf8]{inputenc}

\usepackage{microtype}

\usepackage{inconsolata}

\usepackage{graphicx}

\usepackage{svg}
\usepackage{url}

\usepackage{listings}
\usepackage{xcolor}

\title{Langformers: Unified NLP Pipelines for Language Models}

\author{Rabindra Lamsal\textsuperscript{\textdagger}, Maria Rodriguez Read, Shanika Karunasekera \\
  School of Computing and Information Systems \\
  The University of Melbourne, Australia \\
\textsuperscript{\textdagger}{mail@langformers.com}, {\{maria.read,karus\}@unimelb.edu.au}\\
}

\begin{document}
\maketitle
\begin{abstract}
Transformer-based language models have revolutionized the field of natural language processing (NLP). However, using these models often involves navigating multiple frameworks and tools, as well as writing repetitive boilerplate code. This complexity can discourage non-programmers and beginners, and even slow down prototyping for experienced developers.

To address these challenges, we introduce \textbf{Langformers}, an open-source Python library designed to streamline NLP pipelines through a unified, factory-based interface for large language model (LLM) and masked language model (MLM) tasks. Langformers integrates conversational AI, MLM pretraining, text classification, sentence embedding/reranking, data labelling, semantic search, and knowledge distillation into a cohesive API, supporting popular platforms such as Hugging Face and Ollama. Key innovations include: (1) task-specific factories that abstract training, inference, and deployment complexities; (2) built-in memory and streaming for conversational agents; and (3) lightweight, modular design that prioritizes ease of use.

Documentation: \url{https://langformers.com}

\end{abstract}

\section{Introduction}

Ongoing advancements in transformer-based models \cite{vaswani2017attention} have significantly transformed the field of natural language processing (NLP). Highly capable models continue to emerge, particularly in two key architectures: decoder-only models \cite{brown2020language} and encoder-only models \cite{devlin2019bert}. Decoder-only models (e.g., GPT \cite{brown2020language}, LLaMA \cite{grattafiori2024llama}) are autoregressive, meaning they predict the next word in a sequence, making them highly effective for text generation tasks. In contrast, encoder-only models (e.g., BERT \cite{devlin2019bert}, RoBERTa \cite{liu2019roberta}) work as autoencoders, processing all input tokens simultaneously. This bidirectional understanding makes them ideal for tasks that require rich contextual embeddings.

Although these models have unlocked powerful NLP capabilities, applying them in real-world scenarios remains challenging. Deploying a chatbot powered by a generative large language model (LLM), pretraining or fine-tuning a masked language model (MLM), or setting up semantic search often involves combining various libraries (e.g., transformers \cite{wolf2019huggingface}, LangChain\footnote{https://python.langchain.com/docs/introduction/}, FAISS \cite{douze2024faiss}) and managing infrastructure (e.g., GPU allocation and API endpoints). This complexity can discourage non-programmers and beginners and even slow down prototyping for experienced developers.

To address such challenges, we introduce {\textbf{Langformers}}, an open-source\footnote{Released under Apache 2.0 License} library that consolidates NLP tasks into a single, intuitive API. Langformers preserves direct access to model internals while eliminating boilerplate code. It offers streamlined pipelines for the following NLP tasks: conversational AI, MLM pretraining, text classification, sentence embedding/reranking, data labelling, semantic search, and knowledge distillation.

This paper discusses the library’s architecture and core functionalities, demonstrating its versatility in applications that involve LLMs and MLMs.

\section{Related Work}
Langformers builds upon and integrates several existing frameworks, methodologies, and best practices in NLP, LLMs and MLMs. Below, we discuss key related works that have influenced the design and functionality of the library.

\textbf{Hugging Face Ecosystem}: 
Langformers utilizes several components from Hugging Face’s \textit{transformers} library \cite{wolf2019huggingface}. The Hugging Face ecosystem—including Datasets\footnote{https://huggingface.co/docs/datasets} and Safetensors\footnote{https://huggingface.co/docs/safetensors}—provides a robust foundation for model training and inference. Langformers extends these tools with simplified pipelines for classification, generation, and embedding tasks.

\textbf{MLM Pretraining and Fine-tuning}:
For MLM pretraining, Langformers adopts RoBERTa’s pretraining strategy \cite{liu2019roberta}, abstracting away complexities such as tokenizer training and dataset tokenization. The library offers an end-to-end pipeline for training custom MLMs from scratch.

The library’s text classification functionality is inspired by efficient fine-tuning methods for BERT \cite{devlin2019bert} and RoBERTa \cite{liu2019roberta}. Unlike standalone scripts, Langformers automates dataset preprocessing, label encoding, and training loops while allowing complete customization.

\textbf{Serving LLMs}:
Langformers’ LLM chat interface draws inspiration from services such as ChatGPT\footnote{https://chatgpt.com}, DeepSeek\footnote{https://www.deepseek.com/}, and the Groq LLM Playground\footnote{https://console.groq.com/playground}. Its REST API inference capabilities are influenced by Hugging Face’s Text Generation Inference\footnote{https://huggingface.co/docs/text-generation-inference/en}. Langformers supports built-in memory for conversational agents and real-time streaming.

\textbf{Embedding Models and Semantic Search}:
Langformers integrates Hugging Face models for embedding generation and is compatible with FAISS \cite{douze2024faiss}, ChromaDB\footnote{https://docs.trychroma.com/}, and Pinecone\footnote{https://docs.pinecone.io/guides/get-started/quickstart} for vector search. While libraries like SBERT \cite{reimers2019sentence} and FAISS require manual effort for embedding, indexing, reranking and retrieval, Langformers unifies these components into a single interface.

\textbf{Knowledge Distillation}:
Langformers performs knowledge distillation using techniques from prior work \cite{hinton2015distilling,reimers2020making,lamsal2024semantically,lamsal2025actionable}. The library automates student model configuration and training, significantly reducing boilerplate in traditional distillation pipelines.

\section{Library Design}
Langformers is designed to support LLMs and MLMs across various NLP tasks, with future scalability in mind. This section covers the core design architecture (Section \ref{sec:factory}), available packages (Section \ref{sec:packages}), and options for customizing training, models and tokenizers (Section \ref{sec:train-config}--\ref{sec:tokenizer-config}).

\subsection{Factory Pattern Implementation} \label{sec:factory}

The library adopts the factory design pattern through the \textit{tasks} class, which acts as a centralized interface to instantiate various NLP components. The class has static methods for instantiating task-specific modules such as \textit{generators}, \textit{classifiers}, \textit{embedders}.

\begin{figure*}[htb!]
    \centering
    \includegraphics[width=0.9\textwidth]{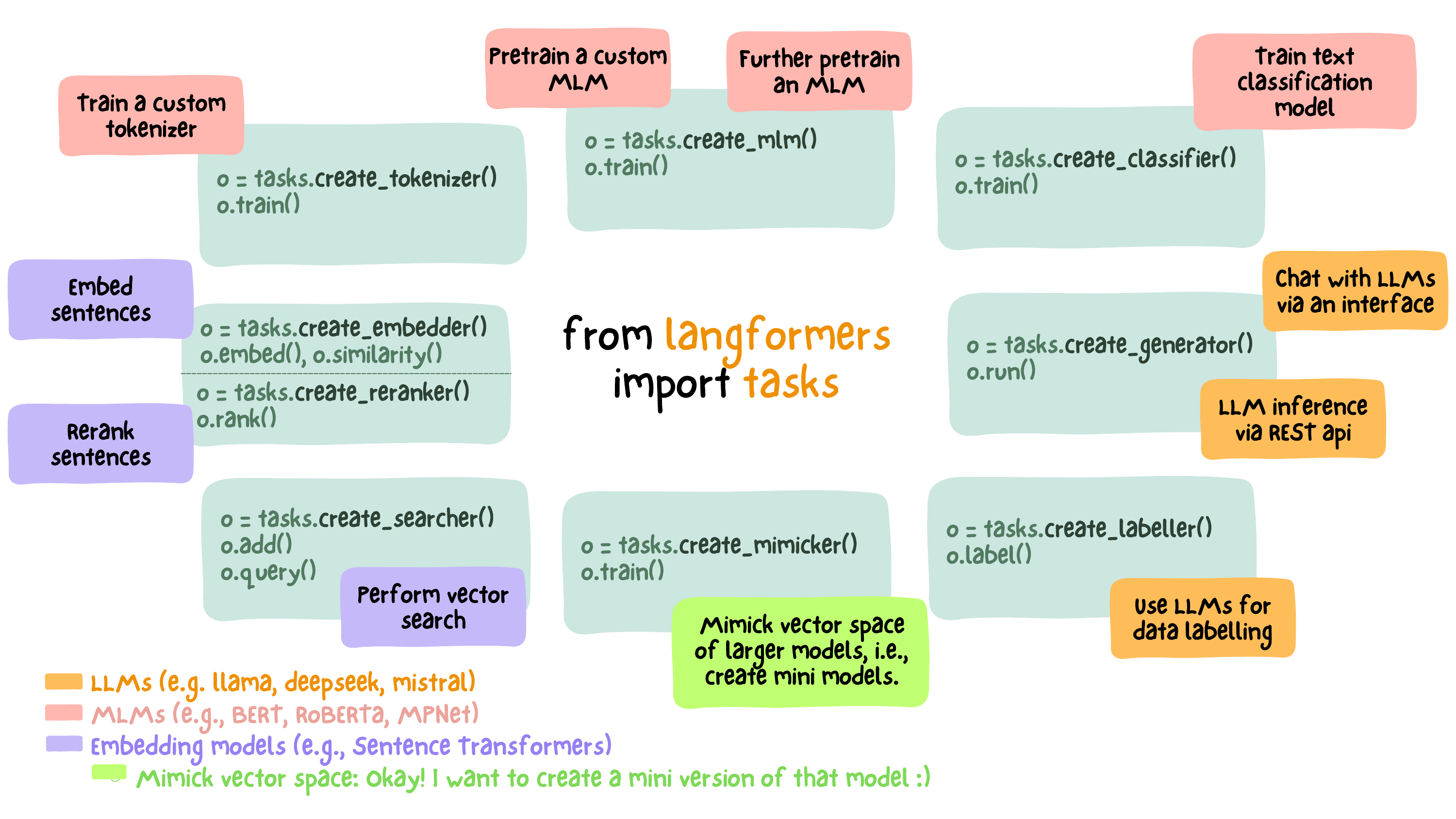}
    \caption{Static methods in the \textit{tasks} class.}
    \label{langformers-tasks}
\end{figure*}

The interface remains uniform across tasks. Example usage:

\begin{lstlisting}
from langformers import tasks

component = tasks.create_<something>() 
component.<do_something>()
\end{lstlisting}

Each \textit{create\_<something>()} method returns a class instance tailored to a specific task, model provider, or database, as defined in the functional packages (Section \ref{sec:packages}). Figure~\ref{langformers-tasks} illustrates the available static methods in the \textit{tasks} class.

\subsection{Packages}
\label{sec:packages}

The library is organized into distinct functional packages, each containing task-specific modules to ensure modularity and extensibility.

\subsubsection{Generators}

The \textit{generators} package provides an interface for interacting with LLMs, supporting both chat-based and REST API deployments. It is compatible with Ollama and Hugging Face models. Modules in this package feature built-in integration with FastAPI\footnote{https://github.com/fastapi/fastapi}, enabling real-time streaming responses.

Key features include:
\begin{itemize}
    \item Memory management for retaining conversation history.
    \item Customizable generation parameters such as temperature, top-$p$ sampling, maximum length, and system prompts.
    \item Authentication hooks integrated with FastAPI dependencies to secure inference endpoints.
\end{itemize}

\begin{figure*}
    \centering
    \includegraphics[width=0.9\linewidth]{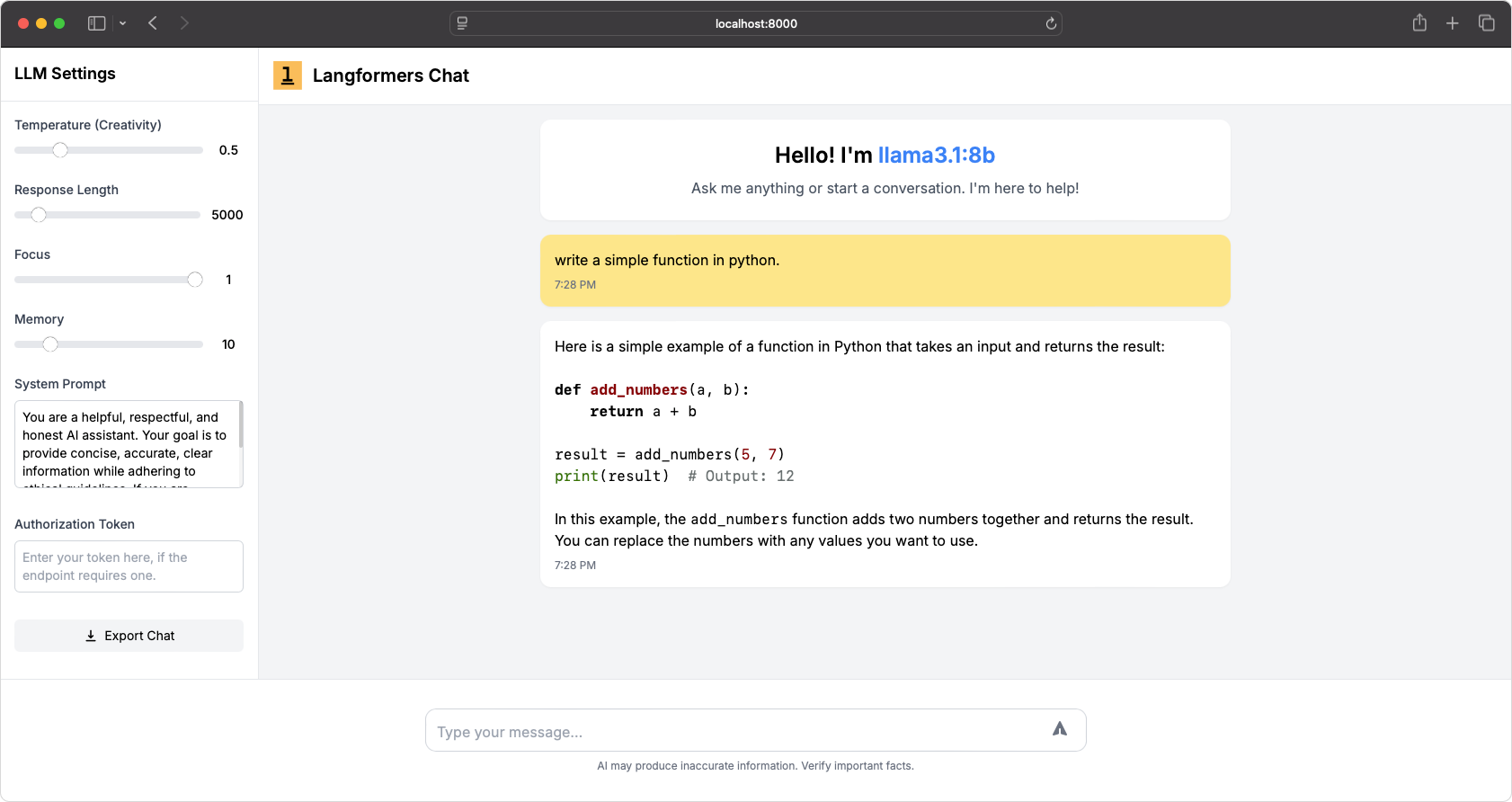}
    \caption{LLM chat interface in Langformers.}
    \label{fig:chat-interface}
\end{figure*}

Modules in this package create an LLM inference endpoint at \textit{host:port/api/generate}. A web-based chat interface (as shown in Figure~\ref{fig:chat-interface}) is available at the specified \textit{host:port}. Alternatively, users can send a POST request to \textit{/api/generate} with following contents:

\begin{itemize}
    \item \textit{system\_prompt}: the system-level instruction
    \item \textit{memory\_k}: the number of previous messages retained
    \item \textit{temperature} and \textit{top\_p}: the hyper-parameters that control how the LLM generates tokens
    \item \textit{max\_length}: the generation limit
    \item \textit{prompt}: the user query
\end{itemize}

\subsubsection{Data Labellers}

The \textit{labellers} package automates text annotation using generative LLMs, reducing manual labelling overhead. It supports both single-label and multi-label classification. Users define labelling conditions as \textit{label (key): condition (value)} pairs, and the model assigns labels accordingly. This functionality is particularly effective for generating weakly-supervised (noisy) labels, categorizing unstructured text, and streamlining annotation pipelines.

\subsubsection{Classifiers}

The \textit{classifiers} package fine-tunes MLMs (e.g., BERT, RoBERTa, MPNet) for tasks such as sentiment analysis, topic classification, and intent detection. Its core module handles end-to-end processing, including dataset preparation, label encoding, model training, and evaluation. It supports CSV input files containing text and label columns, applies tokenization, and performs model fine-tuning using a fully customizable training loop.

The package also includes functionality for loading and using fine-tuned classifiers for inference.

\subsubsection{MLMs}

The \textit{mlms} package enables training custom MLMs from scratch \cite{beltagy2019scibert,lee2020biobert,nguyen2020bertweet,lamsal2024crisistransformers} or further pretraining existing ones on domain-specific data \cite{lamsal2024crisistransformers}. Modules in this package handle tokenizer training, dataset preparation, and MLM pretraining pipelines. Key features include:
\begin{itemize}
    \item Support for custom tokenizers to generate domain-specific vocabularies.
    \item RoBERTa-style pretraining.
    \item Continuing pretraining of existing MLMs.
\end{itemize}

Further pertaining can be extremely useful in low-resource settings, as it has been shown to improve model performance in domain-specific contexts \cite{lamsal2024crisistransformers, lamsal2024semantically}.

\subsubsection{Embedders}

The \textit{embedders} package generates dense vector representations (embeddings) of texts using pre-trained models. It supports popular \textit{sentence-transformers}-like models (e.g., \textit{all-mpnet-base-v2}) \cite{reimers2019sentence} to convert text into embeddings for semantic search and information retrieval. Additional capabilities include:

\begin{itemize}
    \item Cosine similarity computation for tasks such as duplicate detection and content recommendation.
    \item Seamless integration with the \textit{searchers} package for vector indexing and retrieval.
\end{itemize}

\subsubsection{Searchers}

The \textit{searchers} package integrates with vector databases such as FAISS, ChromaDB, and Pinecone to support efficient semantic search across large-scale text corpora. It provides a unified interface for embedding indexing, nearest-neighbor search, and metadata-based filtering.

This package is essential for applications such as document retrieval, question-answering systems, and recommendation engines.

\subsubsection{Rerankers}
Vector search retrieves semantically similar documents, but not necessarily the most relevant ones. 

For example, given the query ``Where is Mount Everest?", a simple retrieval system might rank ``Where is Mount Everest?" above ``Mount Everest is in Nepal." due to surface-level similarity. Reranking addresses this issue by reordering results based on their actual relevance to the query. The \textit{rerankers} package implements this by scoring and sorting documents relative to a given query using Cross Encoders\footnote{https://huggingface.co/models?library=sentence-transformers\&pipeline\_tag=text-ranking}.

Reranking is particularly useful in more advanced pipelines, such as Retrieval-Augmented Generation (RAG) \cite{lewis2020retrieval}.

\subsubsection{Mimickers}

The \textit{mimickers} package implements knowledge distillation by training smaller \textit{student} models to replicate the behaviour of larger \textit{teacher} models. It supports encoder-only architectures for student models and focuses on mimicking embedding space representations. This is particularly beneficial for:
\begin{itemize}
    \item Model compression, reducing inference latency and memory footprint.
    \item Efficient deployment on edge devices or resource-constrained environments.
\end{itemize}

\subsection{Training Configurations}
\label{sec:train-config}

Langformers integrates the \textit{TrainingArguments} class from \textit{transformers}, allowing users to fully customize training pipelines. In addition to the arguments supported by \textit{TrainingArguments}, Langformers includes task-specific parameters to simplify configuration and enhance usability.

For example, a training configuration dictionary may include the key \textit{mlm\_probability}, which controls the likelihood of token masking during MLM pretraining. While this argument is not natively supported by \textit{TrainingArguments}, Langformers separates such task-specific keys and internally assigns them to relevant components. In this case, \textit{mlm\_probability} is passed to the \textit{DataCollatorForLanguageModeling} class in \textit{transformers}. This abstraction enables users to define all training-related parameters in a unified configuration dictionary and streamline the setup when working with complex training pipelines.

Training configurations are currently implemented for the following tasks: training text classifiers, MLM pretraining, and knowledge distillation. Refer to the documentation for a complete list of supported \textit{key: value} pairs for customizing the training processes.

\subsection{Model Configurations}
\label{sec:model-config}

Langformers adopts the RoBERTa architecture \cite{liu2019roberta} as the default backbone for MLM pretraining and model mimicking tasks. This design choice ensures consistent and streamlined training pipelines throughout the library. 

RoBERTa's architecture is both powerful and easy to configure. Langformers utilizes the \textit{RobertaConfig} class from \textit{transformers} to define model hyperparameters. A typical configuration takes the following parameters: 

\begin{itemize}
    \item \textit{vocab\_size}: the tokenizer vocabulary size
    \item \textit{max\_position\_embeddings}: the maximum input sequence length 
    \item \textit{num\_attention\_heads}: the number of attention heads per layer
    \item \textit{num\_hidden\_layers}: the number of encoder layers
    \item \textit{hidden\_size}: the dimensionality of hidden representations
    \item \textit{intermediate\_size}: the size of the feedforward network within each Transformer block
\end{itemize}

\subsection{Tokenizer Configurations}
\label{sec:tokenizer-config}

Langformers supports the creation of tokenizers on custom datasets with the following user-defined configurations:

\begin{itemize}
    \item \textit{max\_length}: the maximum sequence length
    \item \textit{vocab\_size}: the size of the vocabulary
    \item \textit{min\_frequency}: the minimum frequency for a token to be included in the vocabulary
\end{itemize}

\section{Example Use Cases}

Langformers accelerates the development of diverse NLP applications by bridging the gap between cutting-edge NLP research and practical deployment. The following example use cases highlight its adaptability across industries and pipelines.

\subsection{Conversational AI}

\textbf{Seamless Chat Interface:} Langformers enables users to interact with locally hosted LLMs via an intuitive web-based chat interface. This setup combines data privacy with productivity, making it ideal for scenarios such as drafting documents without revealing sensitive data to external cloud APIs.

\noindent \textbf{Intent Classification:} In healthcare settings, Langformers can be used to fine-tune classifiers that automatically categorize patient messages. For example, the system can distinguish between appointment requests and emergency inquiries, improving chatbot triage systems.

\subsection{Content Generation at Scale}

\textbf{API-Driven Inference:} E-commerce platforms can use Langformers to generate thousands of product descriptions on-demand via the REST endpoint. Sparse product specifications can be turned into rich and engaging marketing content.

\noindent \textbf{Sentiment-Aware Labelling:} Product teams can automate the sentiment classification of large-scale customer feedback datasets. The labelling task can replace manual annotation with scalable LLM-powered sentiment analysis.

\subsection{Domain-Specific Language Understanding}

Organizations in highly specialized domains can benefit from custom language modelling capabilities. Legal tech firms can train models on court rulings to capture legal terminology, while biomedical researchers can adapt models such as RoBERTa by further pretraining on PubMed articles. This domain adaptation improves performance on niche vocabulary and specialized downstream tasks.

\subsection{Lightweight Model Deployment}

Langformers supports knowledge distillation pipelines to create smaller student models. Startups can train lightweight models to replicate the semantic behaviour of larger models, significantly reducing memory footprint and inference costs.

\subsection{Semantic Intelligence}

\textbf{Embedding Pipelines:} Recruitment platforms can embed job descriptions and resumes into the same vector space, thus improving candidate-job matching through semantic similarity.

\noindent \textbf{High-Performance Search:} Media platforms can integrate semantic search to present relevant articles to their users based on natural language queries (e.g., ``Climate Change Policies'').

\section{Conclusion}
We introduce Langformers, an open-source library designed to bridge the gap between cutting-edge NLP research and practical deployment, providing a unified and lightweight solution for various language model tasks. With its factory-based design, the library simplifies LLM and MLM pipelines, from pretraining and fine-tuning MLMs to deploying conversational AI. The library is publicly available and open for community collaboration.

Future development will prioritize the continued improvement of core NLP tasks, along with support for additional model providers and model types. In addition, several upcoming features will focus on document preparation for LLM and MLM inference or training.

A complete list of features, up-to-date information, and documentation is available at \url{https://langformers.com}.

\bibliography{custom}

\end{document}